\documentclass[sigconf]{acmart}
\AtBeginDocument{%
  \providecommand\BibTeX{{%
    \normalfont B\kern-0.5em{\scshape i\kern-0.25em b}\kern-0.8em\TeX}}}

\setcopyright{acmlicensed}
\copyrightyear{2024}
\acmYear{2024}

\acmConference[HRI101 Workshop in Human-Robot Interaction '24]{Designing an Intro to HRI Course Workshop at ACM/IEEE International
Conference on Human Robot Interaction (HRI)}{March 11,
  2024}{Boulder, Colorado}
\acmBooktitle{Boulder '24: Designing an Intro to HRI Course Workshop at ACM/IEEE International
Conference on Human Robot Interaction (HRI), March 11--15, 2024, Boulder, CO} 

\begin{document}

\title{HRI in Indian Education: Challenges \& Opportunities}

\author{Chinmaya Mishra}
\email{chinmaya.mishra@mpi.nl}
\orcid{0002-9223-1230}
\affiliation{%
  \institution{Max Planck Institute for Psycholinguistics}
  \city{Nijmegen}
  \country{Netherlands}
}

\author{Anuj Nandanwar}
\email{anuj@ihubiitmandi.in}
\affiliation{%
  \institution{Indian Institute of Technology}
  \city{Mandi}
  \country{India}}

\author{Sashikala Mishra}
\email{sashikala.mishra@sitpune.edu.in}
\affiliation{%
  \institution{Symbiosis Institute of Technology}
  \city{Pune}
  \country{India}
}

\renewcommand{\shortauthors}{Mishra et al.}

\begin{abstract}
    With the recent advancements in the field of robotics and the increased focus on having general-purpose robots widely available to the general public, it has become increasingly necessary to pursue research into Human-robot interaction (HRI). While there have been a lot of works discussing frameworks for teaching HRI in educational institutions with a few institutions already offering courses to students, a consensus on the course content still eludes the field. In this work, we highlight a few challenges and opportunities while designing an HRI course from an Indian perspective. These topics warrant further deliberations as they have a direct impact on the design of HRI courses and wider implications for the entire field. 
\end{abstract}





\maketitle

\section{Introduction}

Recently, there has been a significant increase in efforts to develop advanced general-purpose robots. Companies like Tesla, Fourier Intelligence, and Agility Robotics have already showcased the capabilities of their prototypes and envision starting mass production of their robots soon. Additionally, utility robots such as Roomba\footnote{\href{https://www.irobot.co.uk/en_GB/roomba.html}{Roomba}} and lawnmaster\footnote{\href{https://www.cleva-uk.com/collections/lawnmaster-robot-lawnmowers}{robot lawnmasters}} are becoming increasingly common in households. With an increased presence of robots in our daily lives it is essential to focus on the interactions that these robots have or might have with humans. Human-robot interaction (HRI) is a rapidly growing interdisciplinary field of study focusing on the interactions between robots and humans (design, perception, and evaluation)~\cite{murphy2010human, feil2009human}. The type of interactions and the considerations therein depend on many aspects including the form factor of the robot, its intended application, and the application environment. For example, the interactions one can have with a general-purpose humanoid robot such as Optimus\footnote{\href{https://twitter.com/tesla_optimus}{Tesla Bot}} would be significantly different from the interactions with a Roomba robot. 

To be better equipped for a robot-inclusive society, various aspects of HRI must be thoroughly researched and understood. This includes creating ample opportunities for the next generation to learn about HRI as a field. In recent years a few universities have started offering courses in HRI. Additionally, many studies have discussed and proposed guidelines for HRI courses to be taught in educational institutions~\cite{montebelli2017reframing, zenk2017unconventional}. However, the field has yet to reach a consensus on a standardized curriculum outlining the must-have concepts in an HRI course. In this paper, we discuss the challenges and opportunities from an Indian perspective when designing an introductory HRI course for undergraduate students from all backgrounds (Computer science, Psychology, Cognitive Science, Linguistics, etc). We also suggest a few teaching methods that could be taken into consideration when finalizing an HRI course curriculum. 

\section{Challenges}
In this section, we highlight a few challenges that are inherent when discussing HRI education in India. 

\begin{description}
    \item[Diverse Educational Background]: India has about 65 school education boards as listed by the Ministry of Human Resource Development, India\footnote{\href{https://www.education.gov.in/sites/upload_files/mhrd/files/List_School_Boards.pdf}{List of school education boards}}. Each of these education boards has its curriculum for school students to pursue up to their undergraduate studies. This results in having to cater to a student population that has very diverse educational background knowledge when they enroll at the undergraduate level. 

    This creates a unique challenge when designing coursework for HRI because it becomes difficult to enforce any minimum background knowledge criteria. While the general topics covered by the boards are more or less similar, the focus on the said topics varies significantly between boards. 

    \item[Diverse Languages]: India has more than 121 languages that are categorized as mother tongue, out of which 22 are recognized as scheduled languages\footnote{\href{https://censusindia.gov.in/}{Indian govt. census website}}. Apart from English, many of the education boards allow for the language medium of instruction to be in the mother tongue of the students when possible. Most of the state education boards adopt the state's official language as the medium of instruction. It becomes difficult for students who are not familiar with scientific/ technical terms in English to acclimatize to being taught in the English medium when they join their undergraduate studies. 

    This also compounds the challenge faced in designing coursework for HRI as students need to be made familiar with the keywords and concepts inherent to the field. 

    \item[Cost of Infrastructure]: A key requirement when providing any HRI course is to have the supporting infrastructure available for a hands-on learning experience. This requires significant investments from the institutions. Unfortunately, state-of-the-art robotic systems tend to be very costly (e.g., social robots like Pepper, Ameca, and Furhat) and are out of the reach of most Indian educational institutions. On the other hand, relying on local and affordable robotic systems has the drawback of being unable to provide up-to-date learning opportunities. Ultimately, there is always a trade-off between quality and affordability in Indian institutions, which is very prominent due to the higher costs involved in setting up an HRI lab/ infrastructure. 
\end{description}

\section{Opportunities}
While the diverse educational backgrounds and languages are challenging to navigate, they also offer a unique opportunity for HRI which is a highly multi-disciplinary subject. A key task in teaching HRI to undergraduate students is to introduce them to multi-disciplinary research. With the diverse educational backgrounds of the Indian students, it becomes easier to drive this message. In many ways, a diverse student population as such encourages approaching a problem from various perspectives grounded on the educational background of the students. For example, multilingualism is one of the desired traits in any HRI involving verbal interactions between a robot and a human. It becomes easier for students having varied mother tongues to conceptualize a robotic system that could in theory hold conversations in multiple languages. 

On the other hand, to overcome the barrier associated with the cost of infrastructure the field could come up with better collaboration opportunities and make conscious efforts to foster ties between financially affluent and weak institutions. This could nudge the entire field to become more inclusive and access the untapped potential in regions where researchers do not have access to state-of-the-art facilities. A prerequisite for an effective HRI course in India is the realization of ties with academic and industrial institutions across the world that could facilitate the exchange of various robotic platforms. 

\section{Course Structure \& Teaching Methods}
Given the multi-disciplinary nature of HRI, it is difficult to propose a fixed minimum prerequisite for enrolling in an HRI course at the undergraduate level; in terms of knowledge in engineering, cognitive science, psychology, or other subjects. The \textit{inDia} wheel for HRI education proposed in \cite{montebelli2017reframing} captures the multi-disciplinary aspect of the field quite appropriately. It proposes that HRI education be non-hierarchical and have a mutual relationship among the backbone topics; engineering, natural cognition, artificial cognition, and interaction design.

We propose two key design criteria for an HRI course: \textit{Fully hands-on} course and exposure to robotic systems with \textit{varying form factors}. HRI involves humans interacting with robots with many form factors under a wide range of situations. For example, an assistive robot would involve interactions with a humanoid robot at home, whereas a space robot could involve interactions with an autonomous rover. Moreover, the type of interaction is also defined by the form factor of the robotic system involved. An interaction with a humanoid robot would involve verbal and non-verbal communication with less physical interactions. On the other hand, an industrial robot might require an interface for teleoperation or physical touch. When designing an HRI course, it is therefore crucial to include as many form factors as possible, preferably robotic systems that are very different from each other; for example a Furhat robot for verbal interaction, and an Arduino-controlled wheeled robot for teleoperation. This would expose students to the wide application areas and interaction modalities inherent to HRI. 

Given that the interactions in HRI involve real-world interactions between humans and robots, the most efficient way of teaching HRI is through hands-on experience~\cite{montebelli2017reframing}. A hands-on course structure serves two main functionalities. Firstly, hands-on experience often leads to a better grasp of the fundamentals as students can see the theories in action. Secondly, a hands-on/practical-oriented course design makes it easier to integrate students from different backgrounds as it creates an opportunity for everyone to work towards solving a common problem. HRI is often seen as an engineering-intensive field, which deters students from other disciplines from pursuing it. However, a more inclusive and practical course structure can lead to better results when teaching HRI. A recent work explored two such initiatives to teach non-engineering students to create interactive sequences to be enacted on a robot~\cite{zenk2017unconventional}. It was found that students in both initiatives benefited from the combination of structured education and self-interest-guided explorations. 

To effectively teach HRI to students from diverse disciplines, we propose the following teaching methods:

\begin{itemize}
    \item The course needs to be a combination of structured content and self-explored learning. The fundamental principles need to be introduced gradually while the self-exploration component offers problem statements to students requiring the application of these fundamentals. This would lead to more comprehensive learning. 
    \item The course needs to meet the students mid-way. It is not possible to have a strict prerequisite in various backbone topics (discussed above) for students to enroll in an HRI course. However, this leads to an imbalanced distribution of subject-specific knowledge in the classroom. For example, students from a cognitive science or humanities background would not have the requisite knowledge in programming, whereas students from a computer science background might not have exposure to various theories in cognitive science that are highly relevant to HRI. 

    Thus, the course needs to have two parts. The first part should solely focus on bringing the knowledge base of the classroom up to a common level. The second part should foster the practical application of the acquired knowledge in the form of problem statements. 
    \item During the first few weeks, students should be divided into groups which must comprise at least one student from all the disciplines students come from. Each group would be assigned a fundamental topic from one of the backbone topics and the student(s) with the required background in that topic would be asked to teach the concept to their group mates. 
    \item This needs two tools: an extensive reading list for the students and a guideline on how to break down technical knowledge into simple concepts to be taught to fellow students. Both the reading list and the guidelines need to have ample examples to help students grasp the concept better. 
    \item At the end of the assignment period, students would benefit if they held collective sessions where each group would be tasked with teaching the other students the concepts they have learned. Allowing an environment to freely discuss the subject and questions will help polish the knowledge better. 
    \item The instructor needs to assess the progress and the overall knowledge levels periodically to nudge the students in the right direction. 
    \item Once the structured reading part is over, the students should be offered to choose a problem statement for their hands-on part of the course. These problem statements need to focus on three things: 
    \begin{itemize}
        \item They must cover the fundamental topics.
        \item They must be open-ended enough to allow the students to be more creative when thinking of the solutions. 
        \item They need to cover various aspects of HRI. 
    \end{itemize}
    \item The problem statements should necessitate designing interactions for various robot form factors to cover the various ways HRI can take place. Specifically, students should be exposed to both verbal and non-verbal aspects of HRI.
    \item The final assessment could follow the approach explored in \cite{zenk2017unconventional}, where the students would showcase their design implementations in a public space and answer any questions the audience may have about HRI.
\end{itemize}

\section{Conclusion}
We have tried to highlight some challenges inherent in designing HRI courses from an Indian perspective and also discussed how they could be opportunities to be exploited. These challenges need to be taken into account when deliberating a consensus on HRI course structure. The proposed teaching methods are not only limited to the Indian HRI education system but can be extended to the field in general. An HRI course structure that caters to a diverse pool of students and infrastructure is a much-needed requirement of the field.

\begin{acks}
We would like to thank YaSuei Cheng (Multimodal Language Department, Max Planck Institute for Psycholinguistics, Nijmegen) for sharing her valuable insights on course structuring and teaching methods. 
\end{acks}

\bibliographystyle{ACM-Reference-Format}
\bibliography{paper}

\end{document}